\documentclass[journal]{IEEEtran}
\usepackage{times}
\usepackage{soul}
\usepackage{url}
\usepackage[hidelinks]{hyperref}
\usepackage[utf8]{inputenc}
\usepackage[small]{caption}
\usepackage{graphicx}
\usepackage{amsmath}
\usepackage{booktabs}
\usepackage[ruled]{algorithm2e}
\usepackage{subfigure}
\usepackage{color}
\usepackage{multirow}
\usepackage{array}
\usepackage{algpseudocode}
\usepackage{epsfig}
\usepackage{amssymb}
\usepackage{multirow}
\usepackage{footmisc}
\usepackage{enumerate}
\usepackage{float}

\ifCLASSINFOpdf
\else
\fi
\begin{document}
\title{Multi-scale discriminative Region Discovery for Weakly-Supervised Object Localization}


\author{
        Pei~Lv,
        Haiyu~Yu,
        Junxiao~Xue,
        Junjin~Cheng,
        Lisha~Cui,
        Bing~Zhou,
        Mingliang~Xu,
        and Yi~Yang % <-this % stops a space

%\IEEEcompsocitemizethanks{\IEEEcompsocthanksitem M. Shell was with the Department
%of Electrical and Computer Engineering, Georgia Institute of Technology, Atlanta,
%GA, 30332.\protect\\
%E-mail: see http://www.michaelshell.org/contact.html
%\IEEEcompsocthanksitem J. Doe and J. Doe are with Anonymous University.}% <-this % stops an unwanted space

 \IEEEcompsocitemizethanks{
 \IEEEcompsocthanksitem Pei Lv, Yaiyu Yu, Junxiao Xue, Junjin Cheng, Lisha Cui, Bing Zhou, and Mingliang Xu are with Center for Interdisciplinary Information Science Research, ZhengZhou University, 450000.
 \IEEEcompsocthanksitem Yi Yang  is with Centre for Quantum Computation and Intelligent Systems, University of Technology Sydney, Sydney, NSW, Australia.
 \protect\\
 E-mail: \{ielvpei, xuejx, iebzhou, iexumingliang\} @zzu.edu.cn;  \hfil\break \{yuhaiyu, jj.cheng\}@gs.zzu.edu.cn; \hfil\break yi.yang@uts.edu.au
}% <-this % stops a space
\thanks{}

}

\markboth{IEEE Transactions on Cybernetics,~Vol.~XXX, No.~XXX, August~2019}%
{Shell \MakeLowercase{\textit{et al.}}: Bare Demo of IEEEtran.cls for IEEE Journals}



\maketitle

\begin{abstract}
Localizing objects with weak supervision in an image is a key problem of the research in computer vision community. Many existing Weakly-Supervised Object Localization (WSOL) approaches tackle this problem by estimating the most discriminative regions with feature maps (activation maps) obtained by Deep Convolutional Neural Network, that is, only the objects or parts of them with the most discriminative response will be located. However, the activation maps often display different local maximum responses or relatively weak response when one image contains multiple objects with the same type or small objects. In this paper, we propose a simple yet effective multi-scale discriminative region discovery method to localize not only more integral objects but also as many as possible with only image-level class labels. The gradient weights flowing into different convolutional layers of CNN are taken as the input of our method, which is different from previous methods only considering that of the final convolutional layer. To mine more discriminative regions for the task of object localization, the multiple local maximum from the gradient weight maps are leveraged to generate the localization map with a parallel sliding window. Furthermore,  multi-scale localization maps from different convolutional layers are fused to produce the final result. We evaluate the proposed method with the foundation of VGGnet on the ILSVRC 2016, CUB-200-2011 and PASCAL VOC 2012 datasets. On ILSVRC 2016, the proposed method yields the Top-1 localization error of 48.65\%, which outperforms previous results by 2.75\%. On PASCAL VOC 2012, our approach achieve the highest localization accuracy of 0.43. Even for CUB-200-2011 dataset, our method still achieves competitive results.

\end{abstract}


% \fi
\IEEEpeerreviewmaketitle

\section{Introduction}
Recent years, deep convolutional neural network has been widely used in the tasks of object localization and detection \cite{sermanet2014overfeat, girshick2014rich, girshick2015fast, liu2016ssd, redmon2016you,Xue2018DIOD,Huang2017Semi} through leveraging the vast number of accurate bounding box annotation datasets, such as ILSVRC dataset, which costs a lot of money and manpower. In contrast, weakly supervised learning tries to apply an relatively cheaper and more convenient way by only using image-level supervision \cite{oquab2015object,pinheiro2015image,li2018tell,wei2016hcp,cinbis2017weakly}. Currently, Weakly Supervised Object Localization (WSOL) has drawn increasing attention since it can infer object locations in one given image using only the image-level labels without laborious bounding box annotations for training.

\begin{figure}[!t]
 \centering
 \includegraphics[width=0.95\linewidth]{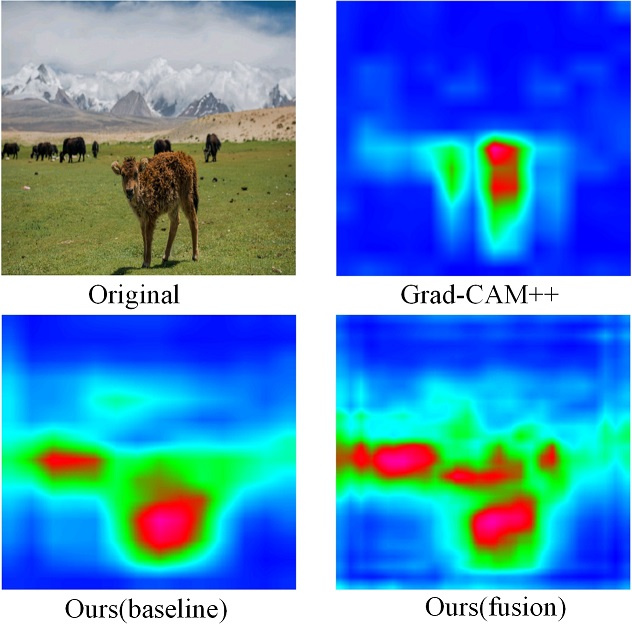}
 \caption{The localization maps generated by our method (baseline and fusion) and state-of-the-art (Grad-CAM++). We adapt a discriminative region discovery to generate the localization map in single layer as the baseline, which is then fused with others to obtain more accurate and integral result. Our method is able to localize more complete regions of multiple instances including small objects.}
 \label{fig:1}
\end{figure}

Zhou \emph{et al.} \cite{zhou2015object} demonstrated that one same network can implement both scene recognition and object localization in a single forward-pass, without having been taught the knowledge of targets. Following \cite{zhou2015object}, some pioneering works \cite{zhou2016learning,zhang2018top,selvaraju2017grad,chattopadhay2018grad,cao2015look,chen2019saliency} have been explored to generate class-discriminative localization maps from pre-trained convolutional classification networks. Zhou \emph{et al.} \cite{zhou2016learning} revisited existing classification networks and proposed Class Activation Maps (CAM) method to locate the regions of interest with only image-level supervision. CAM replaced the top fully connected layer by global average pooling layer to keep the object positions and discovered the spatial distribution of discriminative regions for different classes. However, this method only located a small part of target objects. Meanwhile, it modified the structure of original classification network and had to trade off the model complexity and performance. Selvaraju \emph{et al.} \cite{selvaraju2017grad} introduced a gradient-based way of feature map combination (Grad-CAM) to highlight the discriminative regions, which did not require any modification of the original classification network architecture. But Grad-CAM was unable to localize multiple instances of the same class. Even for those images containing only one single object, Grad-CAM often failed to localize its entire region. In order to overcome above problem, Chattopadhyay \emph{et al.} \cite{chattopadhay2018grad} introduced pixel-wise weighting of the output gradients with respect to a particular spatial position in the final feature map (named as Grad-CAM++). Different from Grad-CAM and Grad-CAM++ built on existing classification network without extra modification, other researchers devised different networks or structures~\cite{singh2017hide,wei2017object,zhang2018adversarial,zhang2018self} to solve the mentioned problems above.

Nevertheless, all these methods inherently ignore the small object with weak response, even no response, in the last convolutional layer estimated by classification network, which cannot be explored by a global average pooling. As pointed out in \cite{singh2018analysis}, large scale variation across object instances, and especially, the challenge of detecting or localizing very small objects stands out as one of the factors influencing the final performance. Moreover, most of existing WSOL methods localize objects from smaller activation maps of the last convolutional layer from classification network, which ignore other meaningful and important information from larger activation maps of different layers.

In this paper, we expect to mine complete regions of multiple objects especially small ones in a simple yet effective way by discovering multi-scale activation maps based on existing classification network without extra modification (as shown in Figure \ref{fig:1}). From the visualization of the activation map of hidden layers of the VGG network pre-trained on ILSVRC 2016 \cite{saleh2016built,bertasius2015deepedge}, we observe the spatial information in the last convolutional layer only focuses on the most discriminative parts of large objects, while other convolutional layers indicate the complementary context of integral objects especially small objects in the form of edges and spatial information~\cite{liu2018receptive}. More importantly, we find that the activation map often exists multiple local maxima especially when multiple similar objects appear at the same time, representing different degrees of response to activated neuron.

Inspired by the above observations, we take advantage of the gradient weight information estimated from multi-scale activation maps to measure the importance of each neuron supporting the prediction by classification network. In detail, we propose a multi-scale discriminative region discovery method to explore the object location information. In order to obtain more integral discriminative regions that spread over the activation map, we apply a discriminative region discovery strategy to find all the local maxima of gradient weights computed by taking partial derivative of particular class score to the activation map. We reformulate the weights of activation maps by taking an average of all the local maxima found in gradient weight maps, and compute a weighted sum of the activation maps in single convolutional layer to obtain corresponding localization maps. Moreover, we fuse the multi-scale localization maps generated from different convolutional layers to produce the final result. Our method is able to deal with the localization problem of incomplete target region with multiple instances of different sizes in the input image by extracting multiple discriminative features from different activation maps.
The key contributions of this paper are summarized as following: \vspace{-0.1cm}
\begin{itemize}
\item We present a weakly supervised object localization method focusing on the localization of complete regions of multiple objects even the small area with weak response rather than the most discriminative ones, which is effective to localize multiple objects in one scenario, especially the small ones.
\item We propose a discriminative region discovery strategy to find the multiple local maxima in gradient weight map and use the new weighted activation maps to locate the regions of interest. Furthermore, activation maps with different scales are fused together to further improve the localization performance. This strategy can be built on any existing classification network without extra modification.
\item The proposed method achieves the new state-of-the-art with the error rate of Top-1 48.92\% on ILSVRC 2016 dataset with only image-level labels, and the localization accuracy of 0.43 on PASCAL VOC 2012 dataset with multi-label image-level labels, which are based on the classification VGGnet.
\end{itemize}

\begin{figure*}[!t]
\centering
\includegraphics[width=0.9\linewidth]{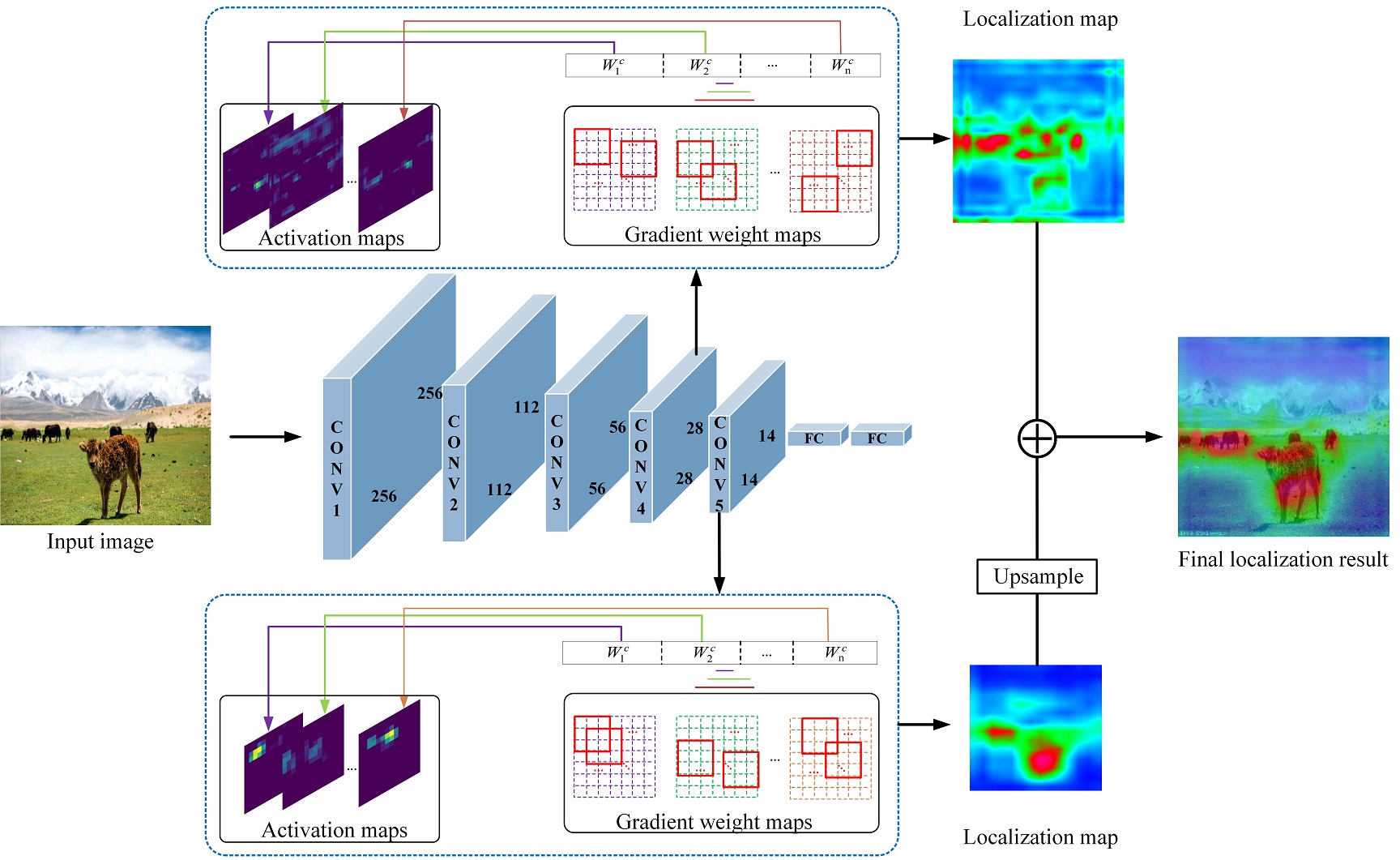}
\caption{The overview of our multi-scale discriminative region discovery method for weakly-supervised object localization. Based on existing classification network, such as VGGnet, we firstly compute the gradient weight maps by taking partial derivative of particular class score to the activation maps. Then a discriminative region discovery strategy is applied to generate corresponding localization maps from different convolutional layers (conv4 and conv5). Finally, multi-scale localization maps are fused into more integral one to locate multiple objects of various sizes. }
\label{fig:2}
\end{figure*}

\section{Related Work}

{\bfseries{Weakly supervised object localization}}.Weakly supervised object localization tries to infer the locations of region of interest in a given image with only image-level supervision. Compared to fully-supervised counterparts \cite{sermanet2014overfeat,girshick2015fast,zhao2018volcano,Wang2017Instance,Li2016Weakly}, WSOL can save huge labor. Zhou \emph{et al.} \cite{zhou2016learning} generated class activation maps by aggregating top-most feature maps using global average pooling layer to keep the object spatial positions. Zhang \emph{et al.} \cite{zhang2018top} proposed an excitation back propagation strategy to generate contrastive attention maps. Selvaraju\emph{et al.} \cite{selvaraju2017grad} put forward a new way of feature map combination  to highlight discriminative regions using the gradient signal without any extra modification of the network architecture. Zeiler and Fergus \emph{et al.} \cite{zeiler2014visualizing} localized various objects in an image by systematically occluding different portions of the input image with a grey square, and dropping the probability of the correct class significantly when the corresponding object is occluded. Singh \emph{et al.} \cite{singh2017hide} presented a random way of hiding patches in a training image to force the network to seek other relevant parts of one integral object with the most discriminative part. Wei \emph{et al.} \cite{wei2017object} introduced one more efficient adversarial erasing approach for localizing more complete object regions progressively. Zhang \emph{et al.} \cite{zhang2018adversarial} applied successfully an adversarial learning strategy which use parallel-classifiers to leverage complementary object regions by erasing its discovered regions from feature maps.

While existing methods are promising, these methods still have several drawbacks. First, the multiple-instance localization of the objects with the same category is not solved very well. Second, it is very difficult to localize objects with different sizes accurately only depending on the feature maps obtained from the last conventional layer, which usually contains very weak responses for these objects.

{\bfseries Multi-instance localization.}
A number of studies formulate the weakly supervised object localization task as a multiple instance learning (MIL) problem \cite{shen2018weakly}. Song \emph{et al.} \cite{song2014learning} developed a framework for learning to localize objects with a smoothed SVM. Cinbis \emph{et al.} \cite{gokberk2014multi} proposed a multi-fold multiple instance learning method which prevents the training process from prematurely locking onto erroneous object locations. Wang \emph{et al.} \cite{wang2015relaxed} come up with a relaxed multiple-instance algorithm to translate multi-instance learning into a convex optimization problem. Oquab \emph{et al.} \cite{oquab2015object} proposed a weakly supervised classification network to predict the location of objects in images. Bency \emph{et al.} \cite{bency2016weakly} proposed a beam searching based approach to detect and localize multiple objects in images.

Although continuous efforts have been made to alleviate the challenges of multi-instance location, there is still one obvious problem in the existing studies. If there are multiple objects of the same kind in an image, the localization algorithms tend to treat them as the same one, particularly for small objects.

{\bfseries Small object detection.}
%Recently some efforts \cite{hariharan2015hypercolumns,he2017mask,lin2017feature,lin2017focal,long2015fully,noh2015learning} have been devoted to improve the performance of small object detection. One effective method is to rebuild the spatial resolution of feature maps by introducing up-convolutions \cite{long2015fully,noh2015learning}. Other methods \cite{hariharan2015hypercolumns,he2017mask,lin2017feature} build a feature-fusion architecture with skip connections to enhance high-level small-scale features. Inspired by these supervised learning methods, we try to involve more activation maps from different convolutional layers to realize better object localization.
In recent years, many approaches~\cite{hariharan2015hypercolumns,he2017mask,lin2017feature,lin2017focal,long2015fully} devoted to promoting the detection accuracy of small object detection. These effective methods aimed to rebuild the spatial resolution of feature maps by utilizing up-convolutions~\cite{long2015fully,noh2015learning}.
\cite{hariharan2015hypercolumns,he2017mask,lin2017feature} further built a bottom-up and top-down structures to strengthen  the features through skip connections. Inspired by these supervised learning methods, we try to involve more activation maps from different convolutional layers to realize better object localization with only image-level labels.

Our work is similar to that proposed by Chattopadhyay \cite{chattopadhay2018grad}. However, their method failed to accurately localize the objects if image contained multiple occurrences of objects with various sizes and classes. In addition, since the feature maps obtained from the last convolutional layer often contain very weak response for small objects, these targets can not be located very well in this way. To deal with these issues which are very common in the real world, in this paper, we present an intuitive multi-scale discriminative region learning method to address the problem of multiple objects localization of the same kind in an image, especially for small objects.

\section{ Multi-scale Discriminative Region Discovery}

The main purpose of our proposed method is to solve the localization problem of multiple objects with the same category or the small objects in images given only image-level labels. To handle above complex situations, a multi-scale discriminative region discovery method is explored (shown in Figure \ref{fig:2}). In this section, we will describe the details of the proposed method.

\subsection{Preparatory knowledge}

\label{prepa}
Our work in this paper is mainly inspired by a series of novel algorithms, namely CAM~\cite{zhou2016learning}, Grad-CAM~\cite{selvaraju2017grad} and Grad-CAM++~\cite{chattopadhay2018grad}. They are based on the fundamental assumption that the final score ${Y^c}$ for a particular class $c$ can be represented as a linear combination of its global average pooled last convolutional layer feature maps ${A^k}$.

\begin{equation}
{Y^c} = \sum\limits_k {w_k^c}  \cdot \sum\limits_i {\sum\limits_j {A_{ij}^k} }
\end{equation}

Each spatial location $(i,j)$ in the class-specific localization map ${L_{ij}}^c$ is then calculated as:

\begin{equation}
L_{ij}^c = \sum\limits_k {w_k^c}  \cdot A_{ij}^k
\end{equation}

$L_{ij}^c$ highly demonstrates the importance of certain spatial location $\left( {i,j} \right)$ for a particular class $c$, thus it can be used as an efficient visual expression of some class predicted by the classification network.

Grad-CAM uses the gradient information flowing into the last convolutional layer of the CNN to understand the importance of each neuron for a decision. It makes the weights $w_k^c$ to be independent of the positions $\left( {i,j} \right)$ of a particular activation map ${A^k}$ by taking a global average pool of the partial derivatives:

\begin{equation}
w_k^c = \frac{1}{Z}\sum\limits_i {\sum\limits_j {\frac{{\partial {Y^c}}}{{\partial A_{ij}^k}}} }
\end{equation}

where $Z$ is the total number of pixels in the activation map, and the final ${Y^c}$ is a differentiable function of the activation maps ${A^k}$. Nevertheless, the unweighted average of partial derivatives leads to the incompleteness of the localization, and the localization map can not cover the entire object, but only bits and parts of it. Grad-CAM++ works around above issue by reformulate the weight $w_k^c$ as:

\begin{equation}
w_k^c = \sum\limits_i {\sum\limits_j {\alpha _{ij}^{kc} \cdot relu\left( {\frac{{\partial {Y^c}}}{{\partial A_{ij}^k}}} \right)} }
\end{equation}

where $relu$ is the Rectified Linear Unit activation function. $\alpha _{ij}^{kc}$ is the gradient weights for a particular class $c$ and activation map $A_{ij}^k$, which is computed by another higher-order derivatives\cite{chattopadhay2018grad}.

Unfortunately, Grad-CAM++ still fails to handle the images containing multiple objects with the same category especially the small object. This is a vital issue as multiple occurrences of objects or small ones in an image is very general in the real scenario.

\subsection{Overview of proposed method}

There will be significant differences among the appearances of objects when looking at them from different perspectives in real world. It means these multiple occurrences of objects will have diverse effects on the classification prediction results especially when they appear in different sizes. For that case, we identify that multiple local maximum exist in activation maps especially when some similar objects appear at the same time, representing different degrees of response to activated neuron since these objects have similar characteristics in some way. In another aspect, the response of small object is weaker compared with the large ones but still stronger than that of local background, the global average of gradient weight will weaken the local maximum response of small objects while enhancing that of most discriminative regions. The basic ideas of the importance of local maximum for the proposed approach are inherent in the understanding of deep convolutional networks.

The overall purpose of our approach is to extract more complete regional context from activation maps to enrich the localization map representation. Figure \ref{fig:2} illustrates the architecture of our method, which uses the gradient weight information from different convolutional layers of neural network as input. First, we take a discriminative region discovery strategy in single convolutional layer to find multiple different local maximum of selected gradient weight maps. These local maximum information are averaged to weight the activation maps, which is different from previous method. These multi-scale localization maps generated by the weighted sum of activation maps contains complementary meaningful information, the smaller one highlights more complete region of large target object, the larger one highlights more details of target objects especially small object. Then localization maps from different layers are fused together to mine more instances even the small ones.

\subsection{Discriminative Region discovery}

\begin{algorithm}[t]
	\caption{Multi-scale discriminative Region Discovery}
	\label{algorithm:1}
	\LinesNumbered
	\KwIn{ Gradient weight map $G^{kc}$ $\left( {\alpha _{ij}^{kc} \in {G^{kc}}} \right)$ of the activation map $A^{k}$ for class $c$.}
	initialize an empty array $l\_m\_arr$ \\
	initialize a maximum filter $Max\_Filter$ with $W \times W$ size \\
	set the stride of the sliding window to $N$ \\
	\While { $G_{ij}^{kc}$ is not traversed}
	{  \eIf{ $G_{ij}^{kc}$ equals to $Max\_Filter(G_{ij}^{kc})$ and $G_{ij}^{kc}$ $>$ 0}
           {
             push back $G_{ij}^{kc}$ into $l\_m\_arr$
           }
           {
             sliding the window with stride of $N$
           }
	}
    $w_k^c \leftarrow average\left( {l\_m\_arr} \right)$ \\
    obtain localization map in single layer $L_{{\rm{conv}}}^c \leftarrow \sum\limits_k {w_k^c \cdot {A^k}}$ \\
    fuse multi-scale localization maps ${L^c} \leftarrow L_{conv4}^c + L_{conv5}^c$

    \KwOut{Localization map ${L^c}$ for class $c$.}
\end{algorithm}

Given an image as the input of one classification network, which can predict classification scores correctly with a single forward-pass, we compute the gradient maps by the back-propagation method mentioned in Section \ref{prepa}. The gradient maps of input image using the class score derivative is the magnitude of the derivative which indicates the influence of each pixel on classification results. We further compute the weight of gradients by higher-order derivatives to obtain the gradient weight maps $G^{kc}$, which will be utilized to measure the importance of each neuron for supporting the network prediction.

Different local maximum are observed when multiple occurrences of objects with various sizes appear in single image. In this situation, a local maximum at location $\left( {i,j} \right)$ for a gradient weight map ${A^k}$ indicates that increasing intensity of pixel $\left( {i,j} \right)$ will have strong discriminative power to the specific filter. These multiple different local maximum indicate other discriminative regions including small object with weak response that are complement to complete feature representation. In order to mine complete regions of target object spreading over the activation map, all local maximum of each gradient weight map are searched, then the weights of activation maps of the convolutional layer are reformulated (as shown in Algorithm ~\ref{algorithm:1}).

In detail, we firstly use $3 \times 3$ window with a stride of 1 to determine whether this pixel is a local maximum, in detail, the values that are equal or greater than all of their 8 surrounding locations as the local maximum, otherwise this value will not be counted. Our method can search all local maximum response spreading over activation maps of different scales in each convolutional layers. This means the number of the local maximum we searched is not fixed according to the different size of sliding window.

The usage of local maximum will capture the importance of activation map without losing the small objects and marginal parts of objects with weak response. In our approach, the flexible parameters of sliding window provide possibilities to achieve better performance in computing weights of activation maps. In particular, our strategy of calculating weights is more effective to explore multiple local maximums response instead of most strong response in single activation map. We use the discriminative region discovery strategy to generate the localization map in the last convolutional layer as the baseline of our method.

\subsection{Multi-scale localization map fusion}

Through the visualization of the activation map from hidden layers of the pre-trained VGGnet on ImageNet, the results show the first two convolutional layers mainly extract image edges. As moving deeper in the network, the convolutional layers extract higher-level semantic features. The third convolutional layer is observed to fire up on prototypical shapes of objects. In particular, the fourth convolutional layer indicates the complementary context of integral objects especially small objects in the form of edges and spatial information, while the spatial information in the fifth convolutional layer only focuses on the most discriminative parts of large objects.

During our experiments, the results of fusing the localization maps generated by the last three convolutional layers indeed remained noisy, which caused inaccurate localization. We found that the fusion of the localization maps generated by the last two convolutional layers are able to achieve the best results. Thus we take full advantage of those two layers to produce more complete region of target object. After the localization maps are generated by estimating the sum of the weight of activation maps at each layer, we further upsample the localization map generated at fifth layer into the same size as the fourth one, then fuse these two localization maps by a simple element wise sum to produce the final localization map. We adapt the multi-scale localization maps of the last two convolutional layers  fusion as the fusion version of our method.

\section{Experiments}

\subsection{Experiment setup}

\textbf{Datasets and evaluation metric} We validate the proposed method on three important public datasets: ILSVRC 2016 \cite{krizhevsky2012imagenet}, CUB-200-2011~\cite{wah2011caltech} and Pascal VOC 2012~\cite{everingham2010pascal}, which are all annotated with image-level labels.

ILSVRC 2016 contains 1.2 million images of 1,000 categories for training and 50,000 images for the validation. CUB-200-2011 has 11,788 images in total of 200 categories with 5,994 images for training and 5,794 images for testing. The localization metric suggested by \cite{russakovsky2015imagenet} is leveraged for comparison, which calculates the percentage of the images whose bounding boxes have over 50\% IoU with the ground-truth.

PASCAL VOC 2012 contains 5,717 images of 20 categories for training. Each image in it has bounding box annotations, and is also multi-label. We use the metric proposed in \cite{everingham2010pascal} to evaluate our approach. This metric defines an Intersection over Union (IoU) $Loc_I^c\left( \delta \right)$, for a particular class $c$, threshold value $\delta$ and input image $I$, which calculates a ratio between the number of non-zero pixels lie inside of bounding box and the sum of non-zero pixels lie outside of bounding box and the area of bounding box.

\footnotetext[1]{These methods are implemented by ourselves on this dataset.}

\textbf{Implementation details} We evaluate the proposed method on the basis of VGGnet \cite{simonyan2014very}. As for ILSVRC and CUB-200-2011 dataset, our classification model is off-the-shelf VGG-16 model from Caffe Model Zoo. We upsample the final localization map into the same size as original input image and scale its values between 0 and 1, which are then fed into threshold segmentation algorithm to generate bounding box at the input image.While as for PASCAL VOC dataset, our classification model is fine-tuned on VOC 2012 based on pre-trained VGG-16 model from the Caffe Model Zoo. We train this network on NVIDIA GeForce GTX 1080 Ti GPU with 11GB memory. Particularly, since its ground-truth including multiple object of various sizes and classes, we measure the localization ability of the proposed method using explanation maps generated by the point-wise multiplication of weighted sum of activation maps and original input image. $\partial $= 0.25 is used as the threshold value. We change pixel value higher than  0.25 to 1.0 in the weighted sum of activation maps. The accuracy value of $Loc^{c}_{I}$ is computed by taking average of $Loc^{c}_{I}$ per label per image.

\subsection{Quantitative comparisons with the state-of-the-arts}

We adapt the multi-scale region discovery method to generate the localization map in the last convolutional layer as the baseline, and further fuse the localization maps generated from multi-scale layers as the fusion version. Table \ref{tab:1} illustrates the localization error on the ILSVRC 2016 validation set. Our baseline method achieves 49.81\% and 37.70\% of Top-1 and Top-5 localization error respectively. The localization errors are further reduced to Top-1 48.65\% and Top-5 34.20\% through fusing the localization maps generated from different-level convolutional layers, and the localization performance is improved by 2.75\% and 5.8\% compared to SPG. We illustrate the results on CUB-200-2011 in Table \ref{tab:2}. Our method achieves 54.86\% and 43.60\% of Top-1 and Top-5 localization error, which are competitive results compared to other state-of-the-arts. In fact, the CUB-200-2011 dataset is a well-designed benchmark for fine-grained object recognition, and a certain amount of annotations are labeled only for the most discriminative region, while the main purpose of our method focuses on discovering multiple objects of various sizes more completely. For the multi-label dataset PASCAL VOC 2012, it also achieves the remarkable performance. Table \ref{tab:3} shows that our method achieves the highest localization accuracy of 0.43 among all those methods. The reason why our method is able to achieve the best results on ILSVRC and Pascal VOC datasets, is that they contain plenty of scenarios where multiple targets of different sizes or small targets exist, the proposed method can explore these cases efficiently.

\begin{table}[!t]
\begin{center}
 \setlength{\tabcolsep}{5mm}{
 \begin{tabular}{l|c|c}
 \hline
  \toprule
  Methods                                 & top-1 error & top-5 error  \\   \hline
  Backprop \cite{simonyan2013deep}        & 61.12       & 51.46        \\
  c-MWP \cite{zhang2018top}               & 70.92       & 63.04        \\
  CAM \cite{zhou2016learning}             & 57.20       & 45.14        \\
  Grad-CAM \cite{selvaraju2017grad}       & 56.51       & 46.41        \\
  Grad-CAM++ \cite{chattopadhay2018grad}  & 52.76       & 40.22        \\
  HaS \cite{singh2017hide}                & 54.53       & -            \\
  AcoL \cite{zhang2018adversarial}        & 54.17       & 40.57        \\
  SPG \cite{zhang2018self}                & 51.40       & 40.00        \\   \hline
  Ours(baseline)                          & 49.81       & 37.70        \\   \hline
  Ours(fusion)                            & 48.65       & 34.20        \\
  \bottomrule
 \end{tabular}}
\end{center}
 \caption{Localization error on ILSVRC validation set.}\label{tab:1}
\end{table}

\begin{table}[!t]
\begin{center}
\setlength{\tabcolsep}{5mm}{
\begin{tabular}[h]{l|c|c}
  \toprule
  Methods                                 & top-1 error & top-5 error  \\ \hline
  CAM\footnotemark[1] \cite{zhou2016learning}             &  62.58      & 58.18        \\
  Grad-CAM\footnotemark[1] \cite{selvaraju2017grad}       &  61.38      & 55.64        \\
  Grad-CAM++\footnotemark[1] \cite{chattopadhay2018grad}  &  60.62      & 49.12        \\
  AcoL \cite{zhang2018adversarial}        &  54.08      & 43.49        \\
  SPG \cite{zhang2018self}                &  53.36      & 40.62        \\  \hline
  Ours(baseline)                          &  55.39      & 45.61        \\  \hline
  Ours(fusion)                            &  54.86      & 43.60        \\
  \bottomrule
\end{tabular}}
\end{center}
 \caption{Localization error on CUB-200-2011 test set. }\label{tab:2}
\end{table}

\begin{table}[!t]
\begin{center}
 \setlength{\tabcolsep}{8mm}{
\begin{tabular}[h]{l|c}
  \toprule
  Methods                                 & $Loc_I^c\ $ \\ \hline
  Grad-CAM \cite{selvaraju2017grad}       & 0.28        \\
  Grad-CAM++ \cite{chattopadhay2018grad}  & 0.38        \\ \hline
  Ours(baseline)                          & 0.39        \\ \hline
  Ours(fusion)                            & 0.43        \\
  \bottomrule
\end{tabular}}
\end{center}
\caption{Localization accuracy on PASCAL VOC 2012 set.}\label{tab:3}
\end{table}

\subsection{Ablation study}

\begin{figure}[!t]
 \centering
 \includegraphics[scale=0.9]{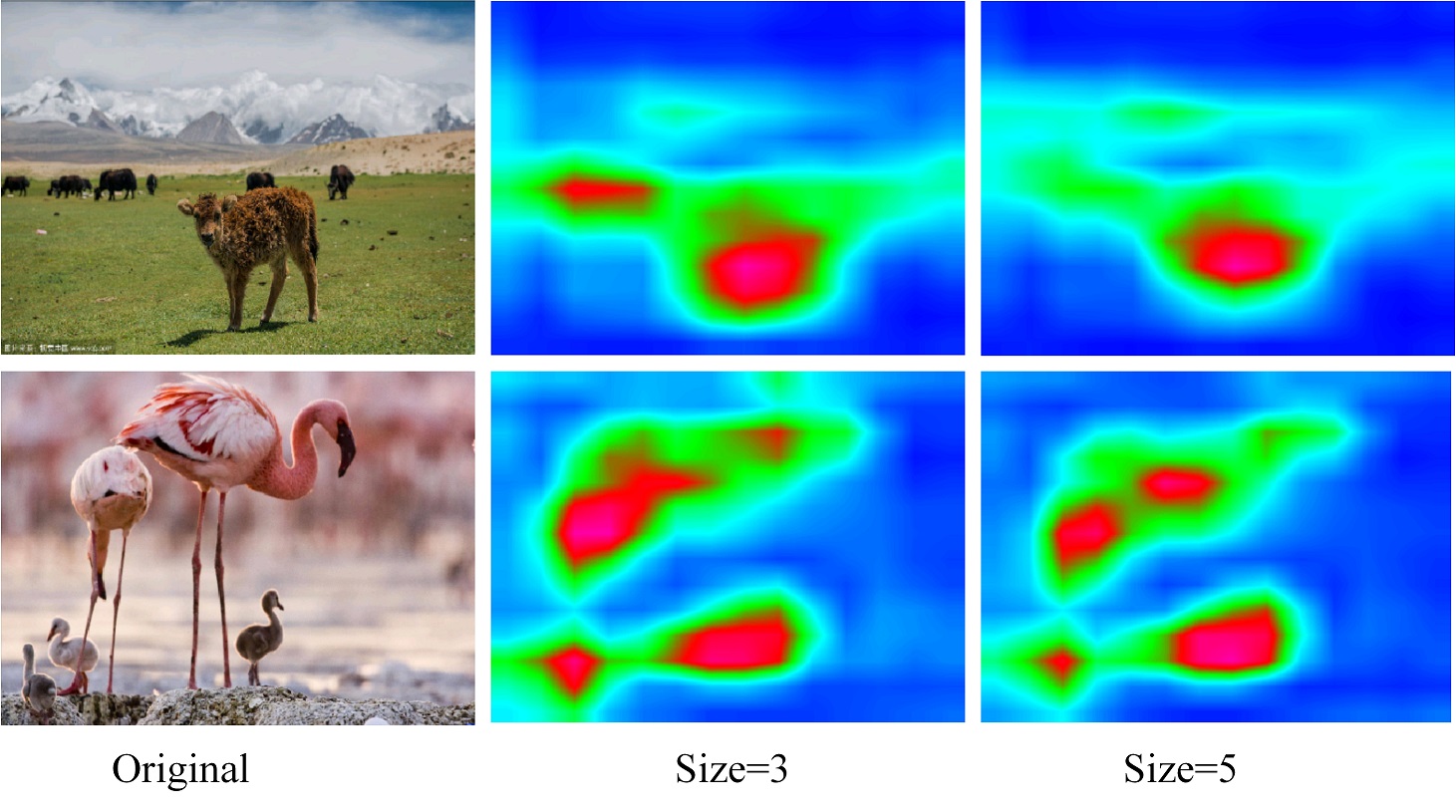}
 \caption{Localization maps generated by different size of sliding window.}
 \label{fig:6}
\end{figure}

In the proposed method, the weights of activation maps are computed via discriminative region discovery strategy. We identify the weights by taking average of the local maximum searched in gradient weight maps, the number of local maximum in each activation map is not fixed. In order to inspect its influence on localization, we discovery the effect of the number of local maximum by testing different size of sliding window to generate various weights of the activation maps from the convolutional layer. The activation maps obtained from the last conventional layer contain very weak response for small objects, as shown in Figure \ref{fig:6}, the smaller sliding window can capture the local maximum produced by small area objects, the larger sliding window will loss them. Our baseline method obtain the best performance in Top-1 error with 49.81\% when the size of sliding window is $3 \times 3$, and the error rates clime up when the sliding window become bigger. This means more local maximum response are searched, the more representation of localization map are explored, our discriminative region discovery strategy is effective to not only smaller activation maps in high-level convolutional layer but also bigger activation maps in fourth convolutional layer.As shown in Table \ref{tab:4}, we can conclude:

\begin{table}[!t]
\begin{center}
\setlength{\tabcolsep}{5mm}{
\begin{tabular}[h]{c|c|c|c}
  \toprule
  layer                  & size & top-1 error & top-5 error \\   \hline
  \multirow{2}{*}{Conv5} & 3    & 49.81       & 37.70       \\   \cline{2-4}
                         & 5    & 52.65       & 39.02       \\   \hline
  \multirow{2}{*}{Conv4} & 3    & 66.74       & 56.74       \\   \cline{2-4}
                         & 5    & 66.83       & 57.24       \\
  \bottomrule
\end{tabular}}
\end{center}
\caption{Localization error with different size of sliding window on ILSVRC 2016.}\label{tab:4}
\end{table}

\begin{table}[!t]
\begin{center}
\scalebox{0.9}{
\begin{tabular}[h]{c|c|c|c|c}
  \toprule
  \multirow{2}{*}{Fusion Layer}&\multicolumn{2}{c|}{ILSVRC 2016}&\multicolumn{2}{c}{CUB-200-2011}\\ \cline{2-5}
                               & top-1 error                    & top-5 error  &top-1 error &top-5 error  \\   \hline
  Conv5                        &   49.81                        &   37.70      & 55.39      & 45.61       \\   \hline
  Conv5+Conv4                  &   48.65                        &   34.20      & 54.86      & 43.60       \\   \hline
  Conv5+Conv4+Conv3            &   60.97                        &   47.23      & 63.84      & 55.22       \\
  \bottomrule
\end{tabular}}
\end{center}
\caption{Localization error with the fusion of multi-scale localization maps.}\label{tab:5}
\end{table}

\begin{table}[!t]
\begin{center}
\setlength{\tabcolsep}{5mm}{
\begin{tabular}[h]{c|c|c}
  \toprule
  threshold      & top 1-error   & top 5-error    \\   \hline
  0.0            & 57.20         &   48.67        \\   \hline
  0.1            & 55.39         &   45.61        \\
  \bottomrule
\end{tabular}}
\end{center}
\caption{Localization error with different thresholds on CUB-200-2011.}\label{tab:6}
\end{table}

\begin{figure}[!t]
 \centering
 \includegraphics[scale=0.9]{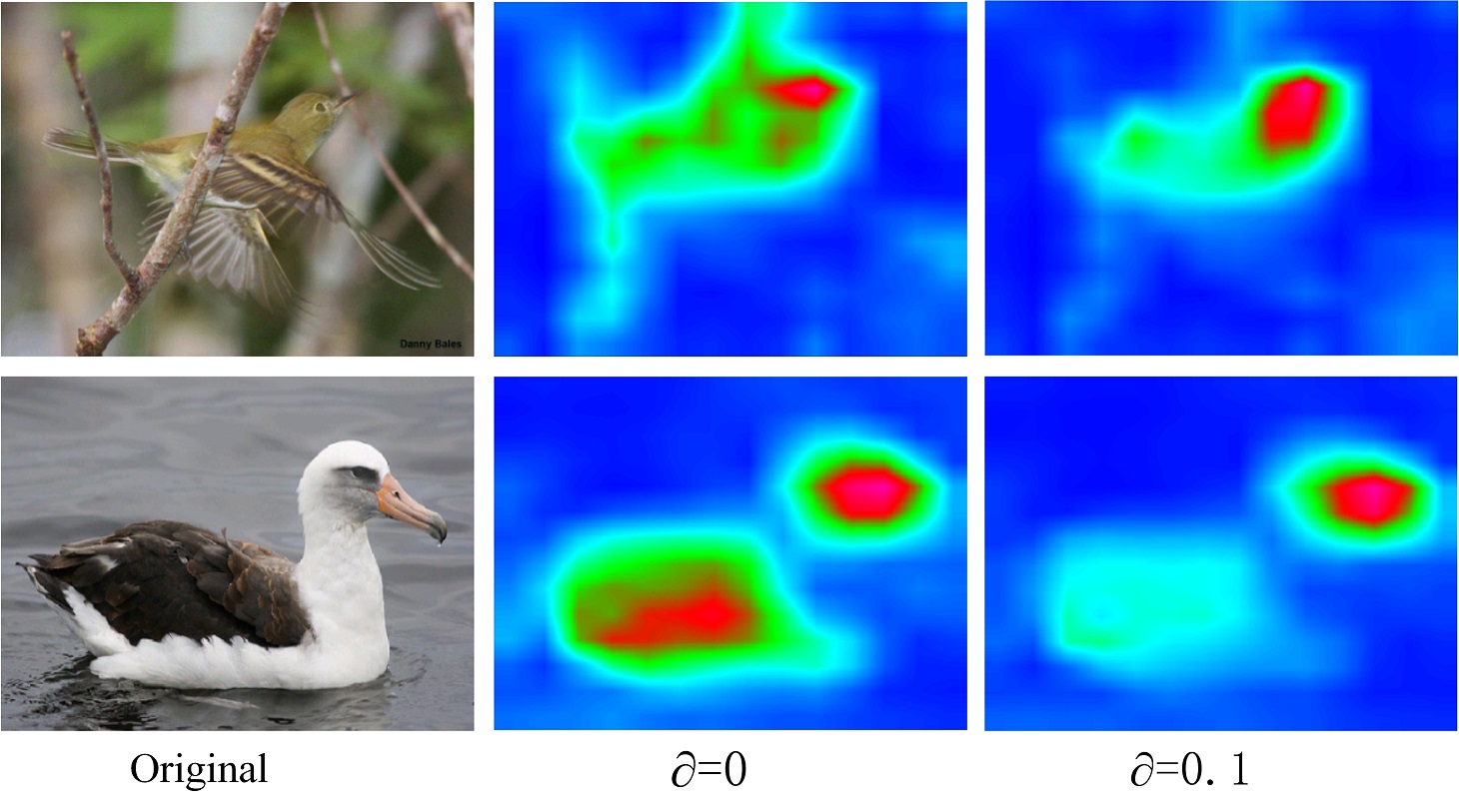}
 \caption{ Localization maps generated by different thresholds of local maximum searching.}
  \label{fig:7}
\end{figure}

\begin{enumerate}
\item The proposed discriminative regions discovery strategy successfully works in computing weights of activation maps, which outperforms the way of global average.
\item The smaller sliding window can improve the performance of the weight calculation method of activation maps. The bigger sliding window will miss some small objects and object edge with weak response.
\end{enumerate}

\begin{figure*}[!t]
 \centering
 \includegraphics[width=0.8\linewidth]{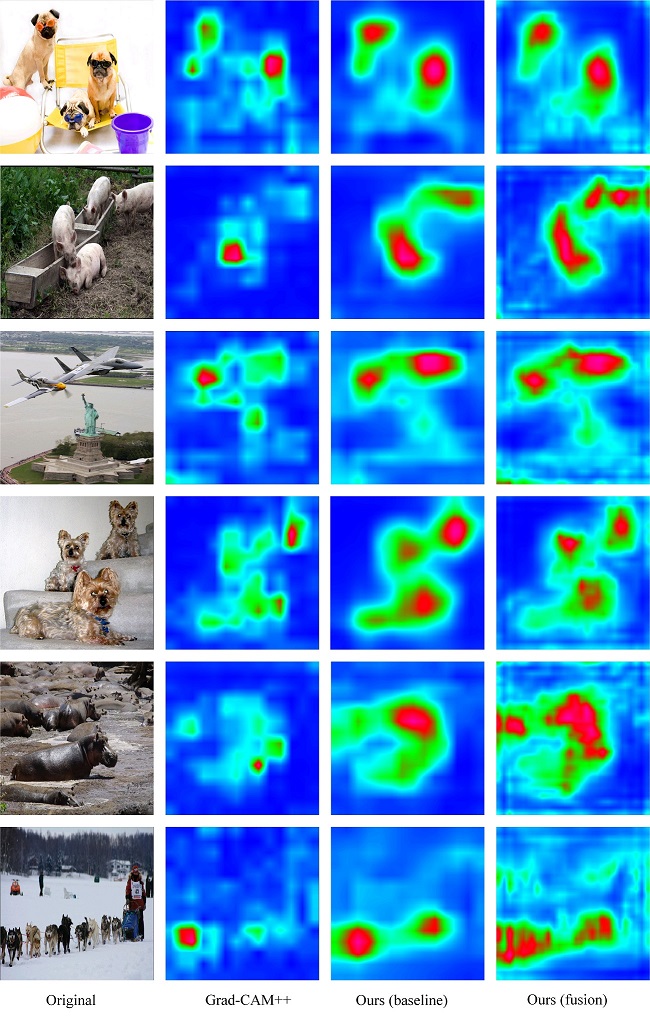}
 \caption{Localization maps of sample images from ILSVRC generated by Grad-CAM++ and our method. Our method can locate more complete regions for multiple objects of different sizes.}
 \label{fig:3}
\end{figure*}

\begin{figure*}[!t]
 \centering
 \includegraphics[width=0.8\linewidth]{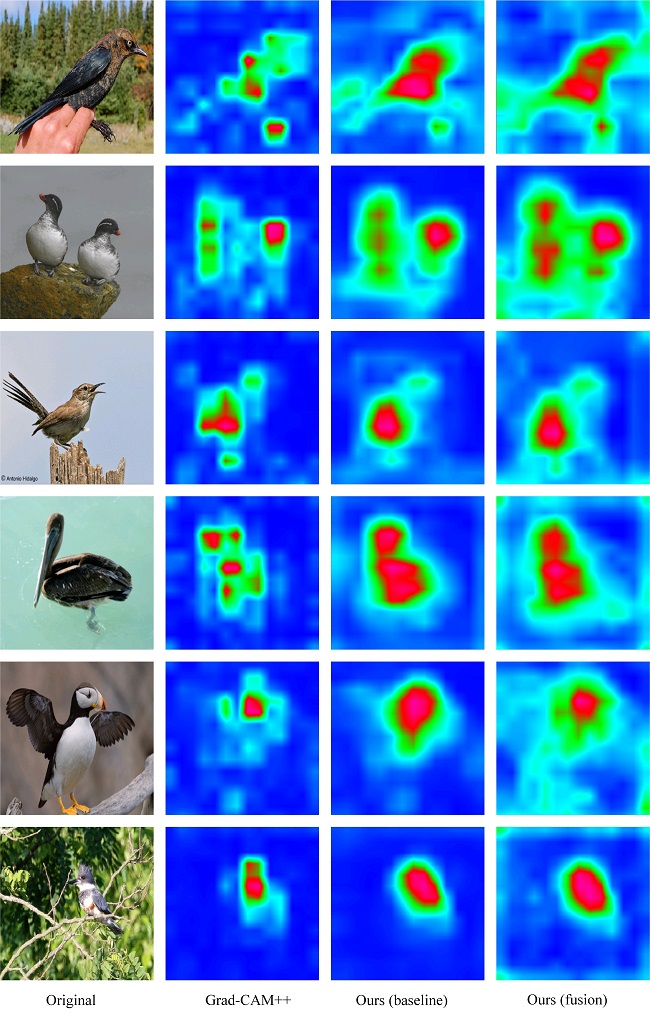}
 \caption{Localization maps of sample images from CUB-200-2011 generated by Grad-CAM++ and our method. Our method can highlight not only the discriminative beak but also colorful body of birds.}
 \label{fig:4}
\end{figure*}

\begin{figure*}[!t]
 \centering
 \includegraphics[width=0.9\linewidth]{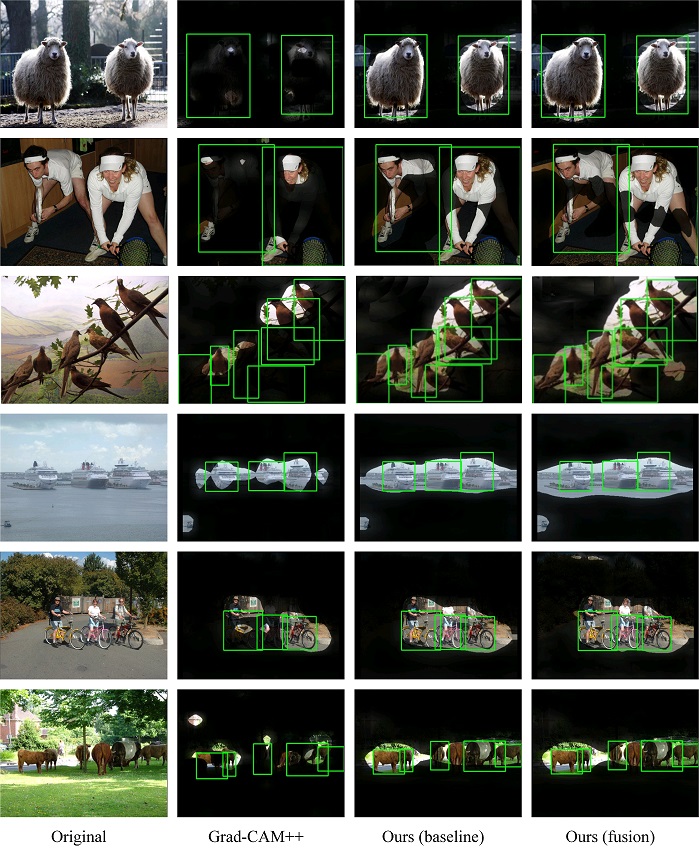}
  \caption{Explanation maps of sample images from PASCAL VOC 2012 generated by Grad-CAM++ and our method. Our method can explore more and accurate target objects in the images. The green bounding box is the ground truth annotations for the images.}
  \label{fig:5}
\end{figure*}

In fact, the fourth convolutional layer actually indicates the location of complete objects, where smaller object features usually present in the form of edges and outlines. Thus we fuse multi-scale localization maps from different convolutional layers to generate the final one. Table \ref{tab:5} give the comparison of localization error among different convolutional layers on ILSVRC 2016 and CUB-200-2011. The fusion version of our method achieves the higher localization accuracy, which indicates we can mine more complete target region when we apply multiple discriminative region discovery strategy in multi-scale layers.

CUB-200-2011 contains 200 categories of similar appearance birds, a large number of its ground-truth locate the most discriminative regions because it is originally designed for fine-grained classification. According to the data distribution, the gradient weight values of the background are 0, and all of them are less than 0.5, thus we set the threshold $\partial $ to 0.1 to capture the most discriminative object regions to obtain better IoU with ground-truth. We take the local maximum whose value is larger than the threshold 0.1 as the final candidate while the searching is implemented in gradient weight maps.Table \ref{tab:6} illustrates our baseline method achieves 55.39\% of Top-1 localization error when we raise the threshold value, which make no compromise on localization error compared to state-of-the-art. Figure \ref{fig:7} shows the visualization results of different thresholds and we can observe the localization map highlights more complete region including the discriminative beak and colorful body of the bird. Moreover, we take a parallel sliding window to search all the local maximum responses in gradient weight maps, thus no additional computational overhead will be introduced.

\subsection{Object localization qualitative results}

Figure \ref{fig:3} shows the localization map of our method and Grad-CAM++. Our localization maps can highlight larger and complete regions of target to generate more precise bounding boxes. For example, in the second , fourth and sixth rows Grad-CAM++ only find a few instances, whereas our method can locate every instance of the same animal. In the first, third and fifth row, our method cover more complete regions of the target objects. Figure \ref{fig:4} displays the localization maps generated by Grad-CAM++ and our approach on CUB-200-2011. Our method successfully locate the complete region such as the discriminative beak and colorful body of the birds. For example, in the first, second and sixth row, our method highlight more complete body than Grad-CAM++. Figure \ref{fig:5} shows the explanation map of the proposed method and Grad-CAM++. The explanation maps generated by our method outperform localization property with more complete target objects in single image. For example, in the first and fifth row, our method highlights more complete regions of sheep and cattle. In the second row, our method discovers more body parts of human instead of only head. Similarly, in the third, fourth and fifth row, every instance of boat, bird and bicycle are explored, while Grad-CAM++ can only find several instances.

We observe the multi-scale discriminative region discovery strategy is successful in discovering the local maximum of gradient weight maps. The weights of activation map via our approach is more effective, which helps obtain the complete region of multiple objects especially the small objects. The localization maps generated by the fourth convolutional layer and the last layer can finally be fused into a robust one, on which the multiple object of different sizes are located better.

\section{Conclusion}

In this paper, we propose a multi-scale discriminative region discovery method for weakly supervised object localization. The proposed method can locate more complete regions of multiple objects in a given image, addressing the problem that the activation maps from the last convolutional layer only contains very weak response for smaller objects. Our proposed method are validated on three important public datasets. The experiments show our discriminative region discovery strategy successfully works on multi-scale convolutional layers by reformulating the weights of activation maps, which greatly improves the completeness of localization representation of multiple objects and outperforms other start-of-the-art localization method based on VGGnet.

Although the proposed method is simple and effective, the localization accuracy still depends on the performance of backbone classification network. It is promising that our method can achieve better result if deeper and better classification network is used, such as the novel ResNet-101. For different networks, more attention should be paid for the choice of different scales of activation maps. Our work can be regarded as a positive attempt. In future work, we will explore the possibility of extending our methods on other weakly supervised tasks of semantic segmentation and object detection

\ifCLASSOPTIONcaptionsoff
  \newpage
\fi

\newpage
\bibliographystyle{IEEEtran}
\bibliography{bare_jrnl}

\clearpage

\begin{IEEEbiography}[{\includegraphics[width=1in,height=1.25in,clip,keepaspectratio]{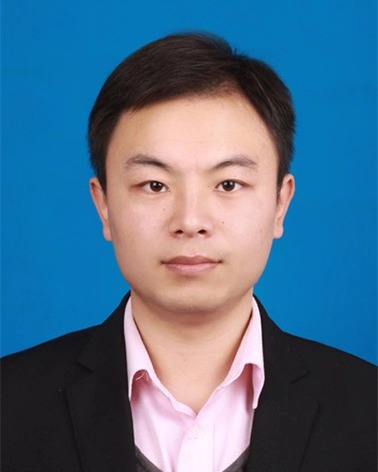}}]{Pei Lv}
is an associate professor in School of Information Engineering, Zhengzhou University, China. His research interests include video analysis and crowd simulation. He received his Ph.D in 2013 from the State Key Lab of CAD\&CG, Zhejiang University, China. He has authored more than 20 journal and conference papers in these areas, including IEEE TIP, IEEE TCSVT, ACM MM, etc.
\end{IEEEbiography}

\begin{IEEEbiography}[{\includegraphics[width=1in,height=1.25in,clip,keepaspectratio]{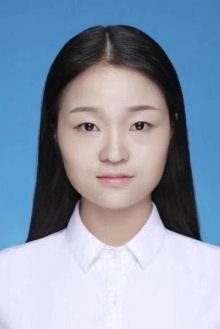}}]{Haiyu Yu}
is a postgraduate student in the School of Software of Zhengzhou University, China. Her research direction is weakly-supervised object localization.
\end{IEEEbiography}

\begin{IEEEbiography}[{\includegraphics[width=1in,height=1.25in,clip,keepaspectratio]{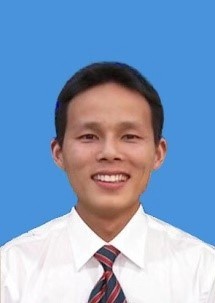}}]{Junxiao Xue}
is an associate professor in the School of Software of Zhengzhou University, China. His research interests include computer vision and computer graphics. He received his Ph.D in 2009 from the School of Mathematical Sciences, Dalian University of Technology, China.
\end{IEEEbiography}

\begin{IEEEbiography}[{\includegraphics[width=1in,height=1.25in,clip,keepaspectratio]{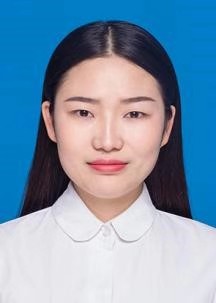}}]{Junjin Cheng}
is a postgraduate student in the School of Software of Zhengzhou University, China. Her research direction is weakly-supervised instance segmentation.
\end{IEEEbiography}

\begin{IEEEbiography}[{\includegraphics[width=1in,height=1.25in,clip,keepaspectratio]{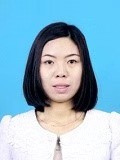}}]{Lisha Cui}
is a Ph.D candidate in the Industrial Technology Research Institute of Zhengzhou University, China, and her research interest is deep learning and computer vision. She got her B.S.degree in computational mathematics from the School of Mathematics and Statistics of Zhengzhou University.
\end{IEEEbiography}

\begin{IEEEbiography}[{\includegraphics[width=1in,height=1.25in,clip]{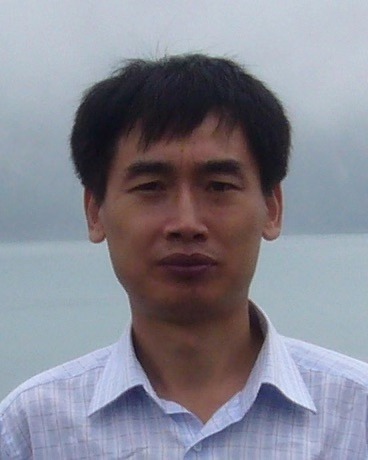}}]{Bing Zhou}
is currently a professor at the School of Information Engineering, Zhengzhou University, Henan, China. He received the B.S. and M.S. degrees from Xi An Jiao Tong University in 1986 and 1989, respectively,and the Ph.D. degree in Beihang University in 2003, all in computer science. His research interests cover video processing and understanding, surveillance, computer vision, multimedia applications.
\end{IEEEbiography}

\begin{IEEEbiography}[{\includegraphics[width=1in,height=1.25in,clip,keepaspectratio]{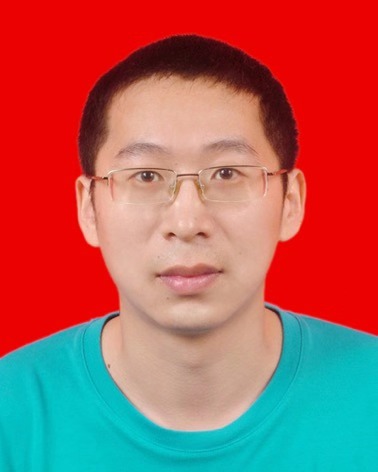}}]{Mingliang Xu}
is a full professor in the School of Information Engineering of Zhengzhou University, China, and currently is the director of CIISR (Center for Interdisciplinary Information Science Research) and the vice General Secretary of ACM SIGAI China. He received his Ph.D. degree in computer science and technology from the State Key Lab of CAD\&CG at Zhejiang University, Hangzhou, China. He previously worked at the department of information science of NSFC (National Natural Science Foundation of China), Mar.2015-Feb.2016. His current research interests include computer graphics, multimedia and artificial intelligence. He has authored more than 60 journal and conference papers in these areas, including ACM TOG, ACM TIST, IEEE TPAMI, IEEE TIP, IEEE TCYB, IEEE TCSVT, ACM SIGGRAPH (Asia), ACM MM, ICCV, etc.
\end{IEEEbiography}

\begin{IEEEbiography}[{\includegraphics[width=1in,height=1.25in,clip,keepaspectratio]{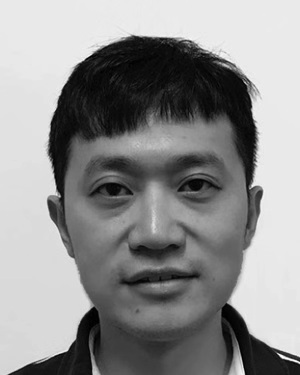}}]{Yi Yang}
received the Ph.D. degree in computer science from Zhejiang University, Hangzhou, China, in 2010. He is currently a Professor with the University of Technology Sydney, Australia. His current research interests include machine learning and its applications to multimedia content analysis and computer vision such as multimedia indexing and retrieval, surveillance video analysis, and video content understanding.

\end{IEEEbiography}

\end{document}